\documentclass[fleqn,10pt]{wlscirep}
\usepackage[utf8]{inputenc}
\usepackage[T1]{fontenc}

\usepackage{graphicx}
\usepackage{multirow}
\usepackage{booktabs}
\usepackage{amsmath}
\usepackage{hyperref}
\usepackage{color, array,amsmath}
\usepackage{algorithm}
\usepackage{algorithmic}
\usepackage{graphicx}
\usepackage{overpic}
\usepackage{graphicx}
\usepackage{multirow}
\usepackage{booktabs}
\usepackage{amsmath}
\usepackage{float}  
\usepackage{lipsum}

\title{Adaptive structure evolution and biologically plausible synaptic plasticity for recurrent spiking neural networks}

\author[1,3,+]{Wenxuan Pan}
\author[1,+]{Feifei Zhao}
\author[1,2,3,4,*]{Yi Zeng}
\author[1,3]{Bing Han}

\affil[1]{Brain-inspired Cognitive Intelligence Lab, Institute of Automation, Chinese Academy of Sciences, Beijing, China.}
\affil[2]{School of Future Technology, University of Chinese Academy of Sciences, Beijing, China.}
\affil[3]{School of Artificial Intelligence, University of Chinese Academy of Sciences, Beijing, China.}
\affil[4]{Center for Excellence in Brain Science and Intelligence Technology, Chinese Academy of Sciences, Shanghai, China.}
\affil[*]{yi.zeng@ia.ac.cn}
\affil[+]{these authors contributed equally to this work}


\begin{abstract}
The architecture design and multi-scale learning principles of the human brain that evolved over hundreds of millions of years are crucial to realizing human-like intelligence. Spiking Neural Network (SNN) based Liquid State Machine (LSM) serves as a suitable architecture to study brain-inspired intelligence because of its brain-inspired structure and the potential for integrating multiple biological principles. Existing researches on LSM focus on different certain perspectives, including high-dimensional encoding or optimization of the liquid layer, network architecture search, and application to hardware devices. There is still a lack of in-depth inspiration from the learning and structural evolution mechanism of the brain. Considering these limitations, this paper presents a novel LSM learning model that integrates adaptive structural evolution and multi-scale biological learning rules. For structural evolution, an adaptive evolvable LSM model is developed to optimize the neural architecture design of liquid layer with separation property. For brain-inspired learning of LSM, we propose a dopamine-modulated Bienenstock-Cooper-Munros (DA-BCM) method that incorporates global long-term dopamine regulation and local trace-based BCM synaptic plasticity. Comparative experimental results on different decision-making tasks show that introducing structural evolution of the liquid layer, and the DA-BCM regulation of the liquid layer and the readout layer could improve the decision-making ability of LSM and flexibly adapt to rule reversal. This work is committed to exploring how evolution can help to design more appropriate network architectures and how multi-scale neuroplasticity principles coordinated to enable the optimization and learning of LSMs for relatively complex decision-making tasks.
\end{abstract}
\begin{document}

\flushbottom
\maketitle
%
%
\thispagestyle{empty}

\noindent Please note: Abbreviations should be introduced at the first mention in the main text – no abbreviations lists. Suggested structure of main text (not enforced) is provided below.

\section*{Introduction}

The brain is a highly heterogeneous and powerful network of tens of billions of neurons possessing unparalleled feats of cognitive functions. Traditional artificial intelligence models are predominantly built on networks with hierarchical feed-forward architectures, different from the highly recurrent connected biological network in the brain~\cite{1}, making it difficult to match the results of natural evolution in terms of function and efficiency. Macro-scale reconstruction studies of human brain structure~\cite{43} confirmed the existence of a large number of non-hierarchical structures in the brain, such as modular structure~\cite{55,56,57,61}, hub structures~\cite{58,59}, small-world structures~\cite{61, 60}. These topological properties enable the brain to better coordinate multiple cognitive functions to adapt to complex and dynamic environments and are also unconventional structures missing in existing brain-inspired AI models.

\begin{figure}[htp]
    \centering
    \includegraphics[width=8cm]{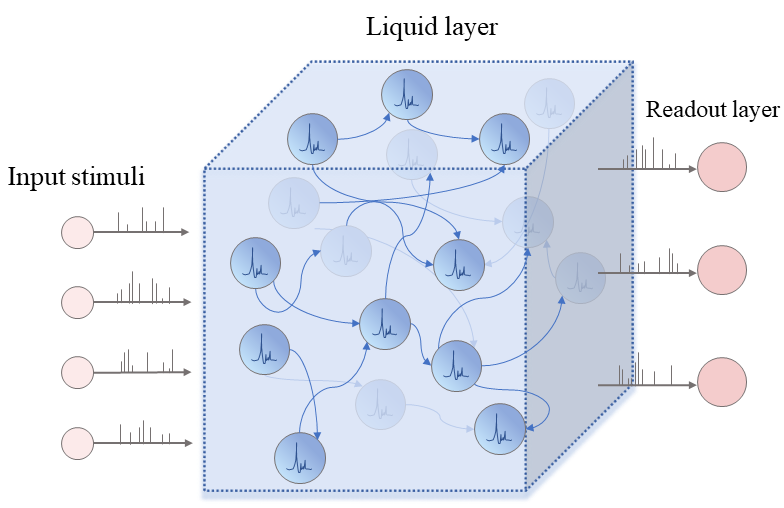}
    \caption{In the traditional definition of the LSM, randomly connected liquid layer neurons receive time-varying signals from external inputs and other nodes. Recursive connection patterns enable input signals to be converted into liquid layer dynamics and then abstracted by the readout layer.}
    \label{fig:lsm}
\end{figure}

Motivated by this, this work focuses on a network structure called Liquid State Machine (LSM) which can generate complex dynamics like the brain and facilitate the processing of real-time tasks. LSM~\cite{26} is a spiking neural network (SNN) structure that belongs to the reservoir, with randomly connected liquid layers and readout layers whose weights can be modified, as shown in Fig.\ref{fig:lsm}. Reservoir computing has achieved some progress in different fields, such as speech recognition \cite{38,39,40}, image recognition \cite{37,41,42}, robot control \cite{29,36}, etc. 

Some LSM models use fixed weights for the liquid layer, probably because its complex recurrent structure is difficult to be trained and optimized, which limits the learning capability and wide application of LSM~\cite{ivanov2021increasing,45,29}. Most of these existing models used gradient-based approach ~\cite{34,31,64,67,35} to train the readout layer without training the liquid layer, resulting in a gap with the real learning mechanism in the brain.  
 Some approaches~\cite{3,25,68,71} tried to train the liquid layer through local synaptic plasticity such as Spike-Timing-Dependent Plasticity (STDP)~\cite{bi1998synaptic} or Hebb~\cite{amit1994correlations}, which is limited to simple tasks. In summary, there is still a need to explore biologically plausible learning rules applicable to LSM to optimize its liquid and readout layers.
 
 In addition, from the structure perspective, the liquid layer is usually fixed after initialization, simply serving as a way of high-dimensional encoding. Some methods~\cite{72,35,68} inspired by deep neural networks superposed multiple LSM layers as a deep LSM to solve machine learning tasks. These approaches have not explored the studies about dynamic LSM' structure search
 in order to adapt to the changing tasks. And in fact, the human brain evolved rather than followed a specific design, which is different from current common AI algorithms.
 Evolution allows the brain's nervous system to be continuously optimized and eventually evolve into non-hierarchical, highly-efficient structures. Inspired by this, some studies~\cite{45,69,34,64,25} proposed evolutionary methods for optimizing the parameters and structures of LSM. ~\cite{29} assessed LSM according to three LSM's properties proposed in ~\cite{26}, and this work encoded the three LSM properties into the chromosome, and optimized the separation property (SP) as the objective function. Using SP as fitness is reasonable because it could reflect the role of evolution in the network dynamic adjustment. However, this work is limited to simple control tasks. ~\cite{34} developed a three-step search method based on the genetic algorithm (GA) to search the network architecture and parameters of LSMs. ~\cite{34,64,25} directly used the experimental data set as a criterion for evaluating the fitness of LSM. These approaches lack effective exploitation of the internal dynamics of LSM.

Considering the various limitations of existing LSM's studies mentioned above, in this paper, we present a brain-inspired LSM with evolutionary architecture and dopamine-modulated Bienenstock-Cooper-Munros (DA-BCM)~\cite{10} Rule. We consider the optimization of LSM from structure and function respectively. \textbf{Structurally}, we optimize the architecture of liquid layer according to an evolutionary perspective to obtain a more brain-inspired effective structure with a higher separation property. \textbf{Functionally}, instead of the gradient-based method, we propose a biologically plausible learning method with the combination of local trace-based BCM \cite{10} synaptic plasticity and global dopamine regulation. The experimental results show that the proposed evolved DA-modulated LSM is able to learn the correct strategy faster and flexibly adapt to rules reversal on multiple reinforcement learning tasks. As reservoir computation exhibits complex dynamics consistent with activity in brain neural circuits, the evolvable LSM based on DA-BCM provides us with a powerful and biologically realistic tool to delve deeper into the learning process of complex networks. This work provides new opportunities for developing more brain-inspired complex and efficient network models, building adaptive learning frameworks, and revealing the evolutionary mechanisms of brain structures and functions.

\section*{Results}

\subsection*{The Evolution and Learning Process}
This paper first evolves the structure of the liquid layer of LSM, and then optimizes the liquid layer and readout layer based on DA-BCM to accomplish online decision making. Evolution randomly initializes $N_{ini}=100$) individuals according to the inputs of different tasks, and then randomly mutates multiple offspring during the mutation process, finally selecting the optimal $N_{opt}=20$ structures for decision making. All experimental results in this work are based on the average of the network structures obtained from multiple random evolution to ensure accuracy and fairness.

Experiments show that evolved individuals are superior to randomly generated models in efficiency. 
Based on the evolved structures, the $N_{opt}$ individuals (agents) are then placed in a specific environment, where the next step action is determined according to the LSM's output. DA-BCM rule dynamically adjusts the strength of LSM's weights through the reward signal fed back by the environment, enabling the agent to learn and survive better in the environment. Model performance is evaluated according to the average cumulative reward of $N_{opt}=20$ individuals within $T$ steps, which is calculated as follows:

\vspace{-0.5cm}
\begin{align}\label{eqreward}
&{}&R&= \frac{{\textstyle \sum_{i}^{N_{opt}}} {\textstyle \sum_{t}^{T}}DA_{t}^{i}}{N_{opt}}
\end{align}

$DA_{t}^{i}$ represents the reward obtained by individual $i$ at step $t$. Therefore, $R$ represents the average reward of all individuals.

\begin{figure*}[t]
\centering  
\includegraphics[width=14cm]{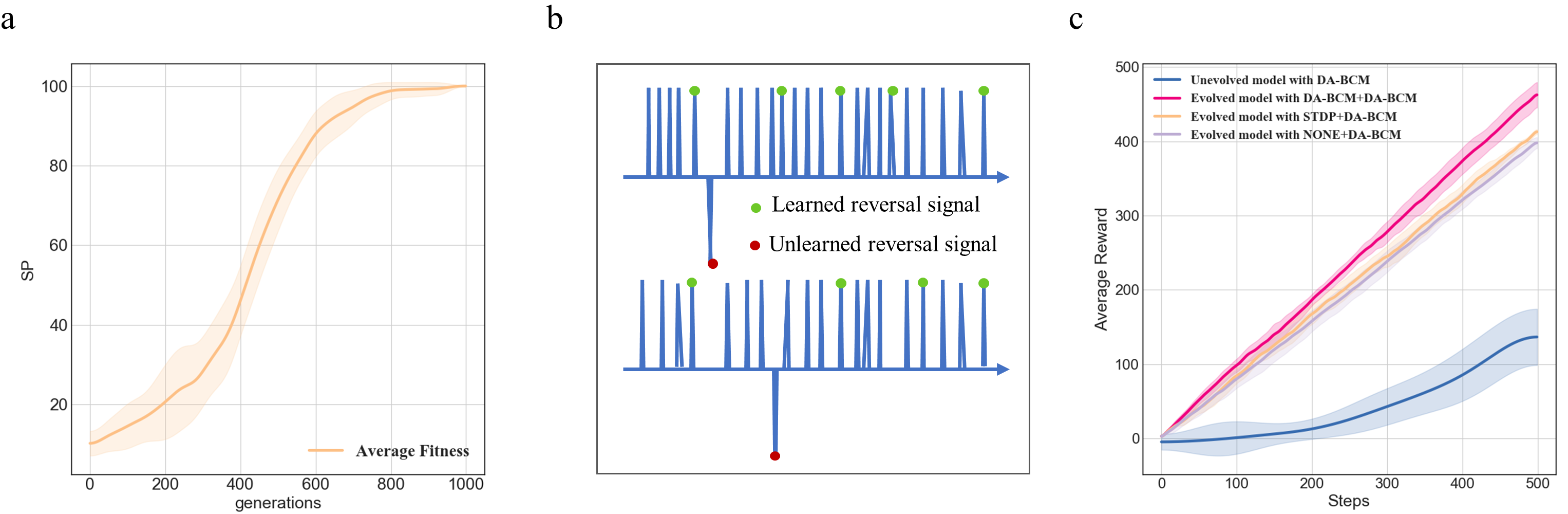}
\includegraphics[width=14cm]{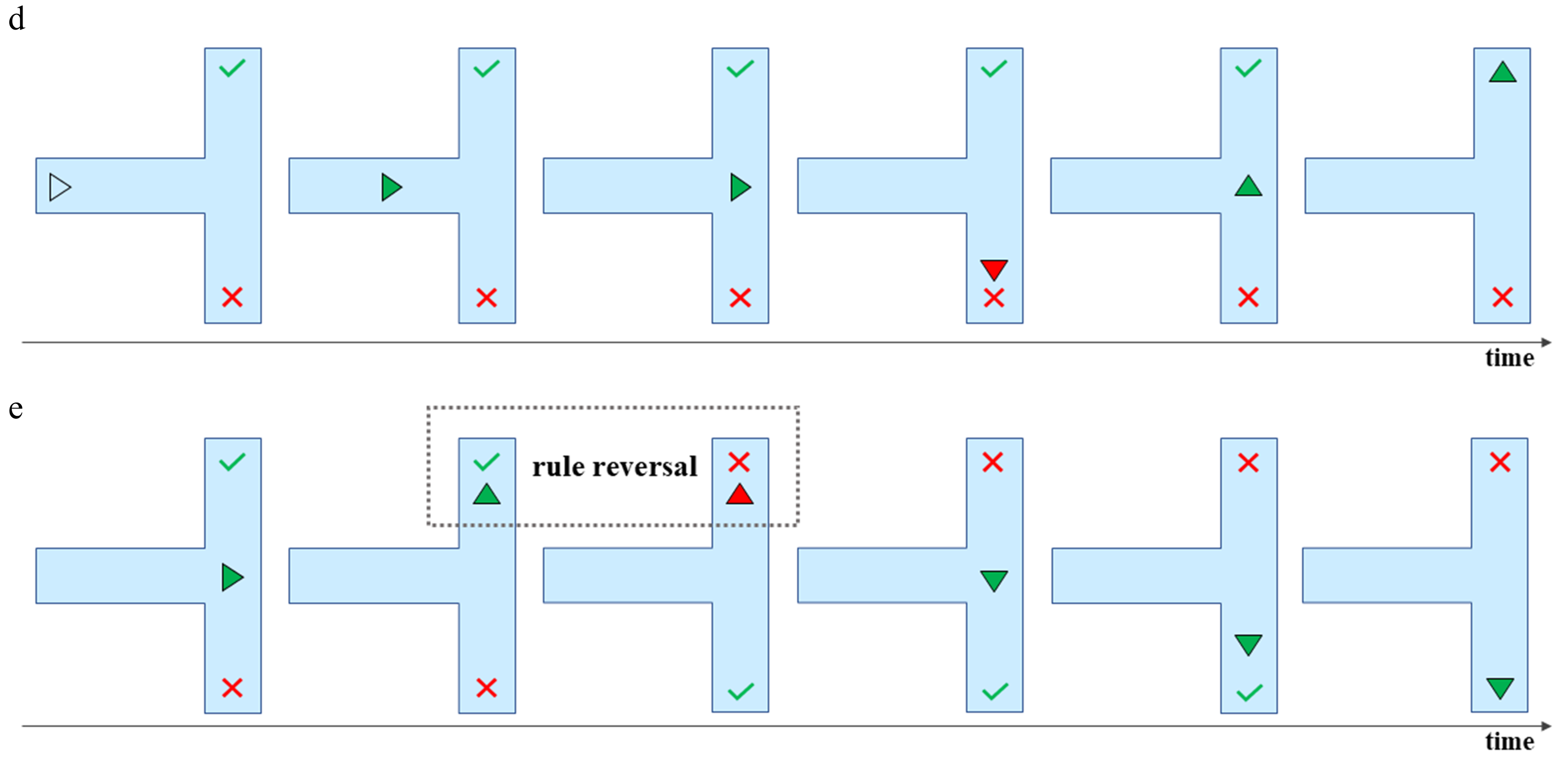}

\caption{Experimental results on T-maze. \textbf{a:} the separation property of evolving LSMs which is calculated from the average of all individuals in population. \textbf{b:} reversal learning results. Green dots indicate that the agent has obtained food, and red dots indicate poison.  \textbf{c:} performance of LSMs (applying different learning rules). Evolved model results are the average performance of $N_{opt}$ individuals, unevolved model results are the average performance of multiple runs. \textbf{d:} the agent learns to change behavior guided by DA regulation.  \textbf{e:} When the rule is reversed, the agent learns to avoid the poison after being punished once.}
\label{fig:rank}
\end{figure*}
\subsection*{T-maze}
\subsubsection*{Experiment Configuration}

We constructed a T-shaped maze (two ends of the maze are food and poison, respectively). Three input neurons representing the agent's observations in three directions (maybe walls, roads, food, poison) feed information into the evolved LSM, and the agent performs actions (forward, left, right) based on the output. The distance difference between the agent and the food before and after executing the behavior is defined as $dis_{m}$, then the reward function for the T-maze task is:

\vspace{-0.5cm}
\begin{align}
\label{eq}
&{}&DA= \begin{cases}
 3,&get\ food\\
-3,&get\  poison\\
1,&dis_{m}<0\\
-1,&dis_{m}\geq 0\\
\end{cases}
\end{align}

An energy mechanism is set up to prevent the agent from wandering or standing still (i.e.hitting the wall all the time) in the maze. Each round is counted from the starting point to the endpoint, where learning time is limited to a certain number of steps, after which the exploration process will be restarted. When the agent receives positive rewards continuously, the positive and negative rewards in the maze will exchange positions with a certain probability, thereby verifying the ability of the model to adapt to reversal learning.

\subsubsection*{Results on T-maze}
The fitness change of the evolved $N_{opt}$ individuals is shown in Fig.\ref{fig:rank}a, which can gradually evolve to reach the maximum value, verifying the evolutionary algorithm's effectiveness. 
Comparing evolved and unevolved models in Fig.\ref{fig:rank}c, we could find that structure evolution improves the learning efficiency and performance of LSM models. Models' performance is calculated using the average reward value of $N_{opt}$ individuals over a period of time $T=500$.
Fig.\ref{fig:rank}d shows how evolved LSMs with DA-BCM learning rule help the agent to find where the food is. Along the way, reward signals of environmental feedback (shown in green and red, respectively) guide agent behavior through dopamine modulation.

\textbf{Reversal Learning. } During the learning process, the agent showed the ability to flexibly adapt to the reversion of the rules, as shown in Fig.\ref{fig:rank}b: after taking the poison for the first time and being punished, the agent can avoid the poison no matter how the positions of the poison and food are changed, which means that agent has the ability to reversal learning and can flexibly handle changes in the environment. Simulation results shown in Fig.\ref{fig:rank}e indicate that the agent exhibits the ability of reversal learning.

\textbf{Ablation Analysis. } Ablation experiments further evaluate the effect of DA-BCM learning rules by applying STDP and DA-BCM to the liquid and readout layers to explore the effect of different learning rules on LSM performance. As shown in Table \ref{tab:tmaze} and Fig.\ref{fig:rank}c, the evolved LSM with liquid and readout layers trained by DA-BCM achieves the best performance and significantly outperforms other models. The worst among all methods is the evolved LSM trained with unsupervised STDP, indicating that the model without any environmental feedback to guide LSM dynamics cannot gain knowledge, causing the average reward $R$ to fluctuate over time with no cumulative trend. Besides, the none+DA-BCM and STDP+DA-BCM (the front of "+" represents the liquid layer learning rule, and the back represents the readout layer learning rule) models achieve similar good performance, indicating that the regulation of DA-BCM at the readout layer can help the model to learn the rules of the environment. The STDP+DA-BCM and DA-BCM+DA-BCM are superior to the none+DA-BCM, which indicates that optimizing the weights of liquid layer is more effective than fixing their weights. Further, the outstanding advantage of DA-BCM+DA-BCM illustrates that our proposed biologically plausible DA-BCM training method can outperform untrained or STDP-trained models, helping to evolve LSM's structure and learn the environment information more efficiently.

\begin{table}[tbp]
  \centering
  \caption{Results of ablation experiments on T-maze.}
    \begin{tabular}{cccc}
    \toprule

    Structure & Liquid Layer & Readout Layer & Performance \\
    \midrule
    Evolved & STDP  & STDP  & -0.9$\pm$5.54\\
    Unevolved & DA-BCM & DA-BCM & 162.0$\pm$12.44  \\
    Evolved & none  & DA-BCM & 399.44$\pm$5.99\\
    Evolved & STDP  & DA-BCM & 414.75$\pm$1.41 \\
    \textbf{Evolved} & \textbf{DA-BCM} & \textbf{DA-BCM} & \textbf{464.4$\pm$5.15} \\
    \bottomrule
    \end{tabular}%
  \label{tab:tmaze}%
\end{table}%

\begin{table}[htbp]
  \centering
  \caption{Reward function of Flappy Bird.}
    \begin{tabular}{cccc}
    \toprule
    \multirow{2}[2]{*}{current state} & \multicolumn{2}{c}{last state $=$ current state} & \multirow{2}[2]{*}{last state $\ne$ current state} \\
          & $dis_f<0$ & $dis_f \ge 0$ &  \\
    \midrule
    0 or 1 & 6     & 6     & 6 \\
    2 or 3 & 3     & -5    & -3 \\
    4 or 5 & 3     & -8    & -5 \\
    6 or 7 & 3     & -3    & -3 \\
    8     & -100  & -100  & -100 \\
    \bottomrule
    \end{tabular}%
  \label{t2}%
\end{table}%

\subsection*{Flappy Bird}
\subsubsection*{Experiment Configuration}
Flappy Bird is a game in which the player controls the bird to move between the gaps of many pipes without colliding as much as possible. The settings of state space and action space are shown in the Fig.\ref{fig:bird}a, the size of action space is 2 (i.e. up or down). We divide the state space into 9 parts according to the positional relationship between the bird and the pipe. The state of the bird and its action at the next moment are the input and output of the evolved LSM. The positive reward is used to encourage the bird to pass the pipes (i.e. the gap between the upper and lower pipes), and the negative reward is used to punish the bird for staying away from the exit of the pipes. The reward is learned by LSM for the adjustment of synaptic strength (based on DA-BCM). When the bird collides with the pipe, the game is over.

Table \ref{t2} illustrates the definition of reward function ($DA$) for Flappy Bird. The reward is determined according to current state and last state, and the distance difference $dis_f$ between the bird and the center of pipe before and after executing a selected behavior. The maximum positive reward is given when the bird is in state 0 or 1. A slightly smaller positive reward is used to encourage shorter distances ($dis_f<0$) to the target location (i.e. empty space between pipes). Correspondingly, if the distance becomes longer ($dis_f \ge 0$), a negative reward is given. The largest negative reward is used to punish hitting the pipe. Models' performance is calculated using the average reward value of $N_{opt}$ individuals over a period of time steps $T=2000$.

\subsubsection*{Results on Flappy Bird}
\begin{figure*}[t]
\centering  
\includegraphics[width=14cm]{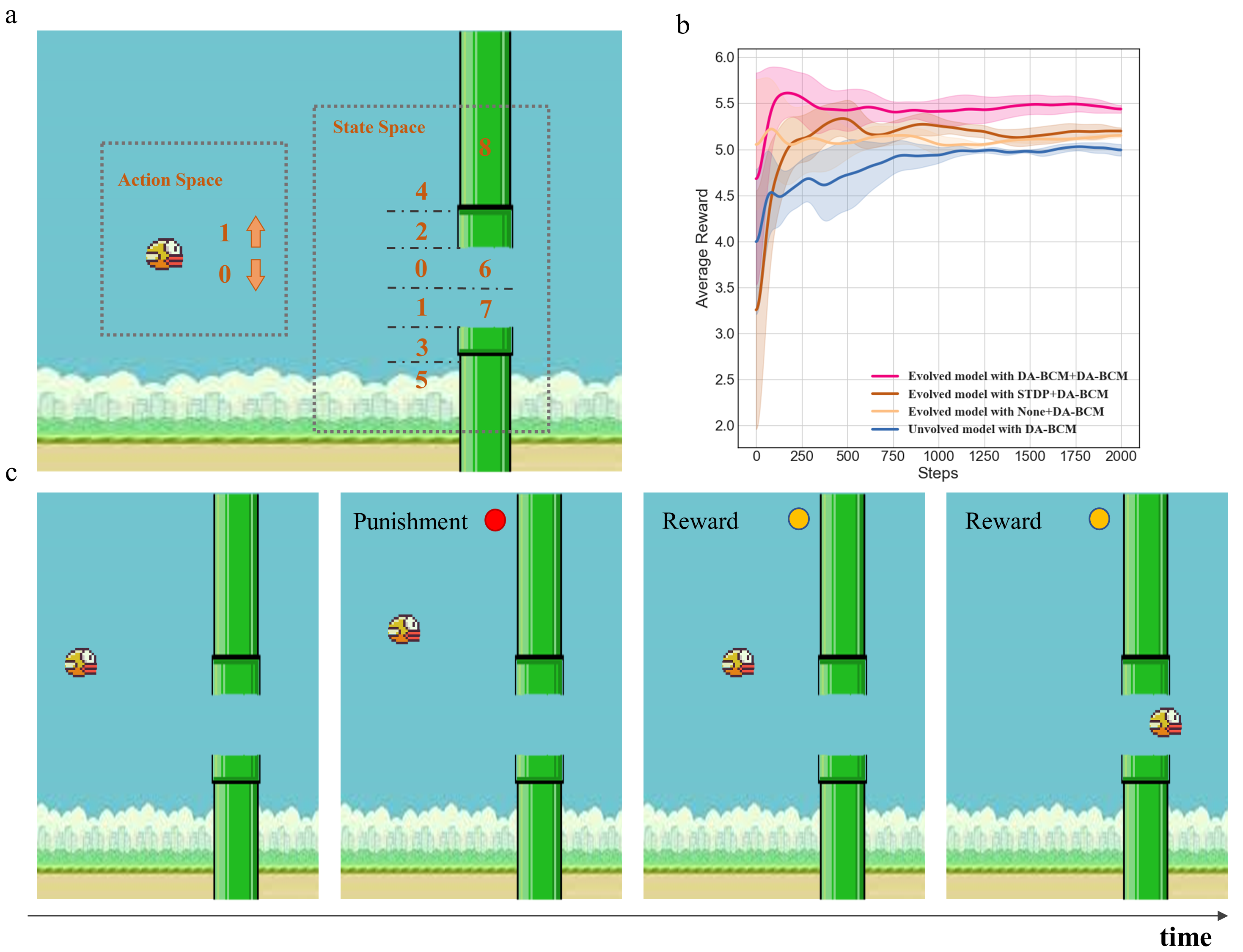}
\caption{Experimental results on Flappy Bird. \textbf{a:} the setup of state space and action space. The whole space is divided into 9 states, where 6 and 7 are the ultimate goals to be achieved.  \textbf{b:} the final performance of all models in the Flappy Bird environment. Evolved model results are the average performance of $N_{opt}$ individuals, unevolved model results are the average performance of multiple runs. \textbf{c:} agents avoid mistakes under the guidance of reward signals.}
\label{fig:bird}
\end{figure*}

\begin{table}[ht]
  \centering
  \caption{Results of ablation experiments on Flappy Bird.}
    \begin{tabular}{cccc}
    \toprule
    Structure & Liquid Layer & Readout Layer & Performance \\
    \midrule
    Evolved & STDP  & STDP  &-54.77$\pm$59.7\\
    Unevolved & DA-BCM & DA-BCM & 4.97$\pm$0.04 \\
    Evolved & none  & DA-BCM & 5.29$\pm$0.00 \\
    Evolved & STDP  & DA-BCM & 5.36$\pm$0.02 \\
    \textbf{Evolved} & \textbf{DA-BCM} & \textbf{DA-BCM} & \textbf{5.43$\pm$0.03} \\
    \bottomrule
    \end{tabular}%
  \label{tab:bird}%
\end{table}%
To verify the validity of the proposed model, we compared the unevolved LSM with liquid layer and the readout layer both trained by DA-BCM (Unevolved model with DA-BCM in Fig.\ref{fig:bird}b), the evolved LSM with non-trained liquid layer and DA-BCM trained readout layer (Evolved model with None+DA-BCM in Fig.\ref{fig:bird}b), the evolved LSM with STDP-trained liquid layer and DA-BCM trained readout layer (Evolved model with STDP+DA-BCM in Fig.\ref{fig:bird}b), and the evolved LSM with DA-BCM trained liquid layer and DA-BCM trained readout layer (Evolved model with DA-BCM+DA-BCM in Fig.\ref{fig:bird}b), respectively. Fig.\ref{fig:bird}b depicts the average reward curves ( Gaussian smoothed) for different models. It is obvious that evolved model with DA-BCM+DA-BCM achieves the best results, while unevolved method is inferior to the evolved methods. Comparing the optimization methods for liquid layer, STDP slightly outperforms the untrained method, while DA-BCM can further bring improvements. Fig.\ref{fig:bird}c shows that our proposed model can guide the bird to fly smoothly through the pipe via dopamine modulation. 

The detailed final performances (average reward $R$ and its variance) of different methods are listed in Table \ref{tab:bird}. LSM trained only with STDP could not finish this task, failing to get positive feedback from the environment, causing the bird to hit the pipe from the start and the game to stop. Each component of the proposed model such as evolution, DA-BCM in the liquid layer and readout layer enables the LSM network to learn faster and better. Thus, we can conclude that our work brings outstanding superiority to optimizing the LSM from the structural and functional perspectives.

\begin{figure*}[ht]
\centering  
\includegraphics[width=7cm]{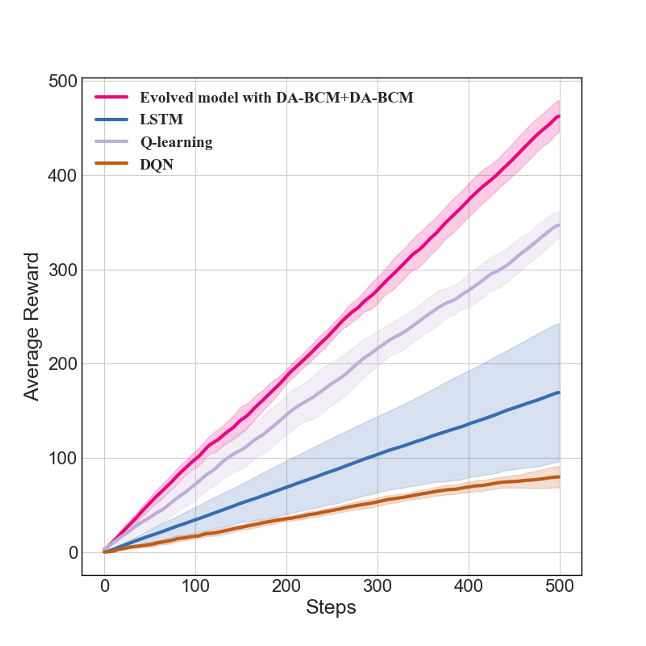}
\includegraphics[width=7cm]{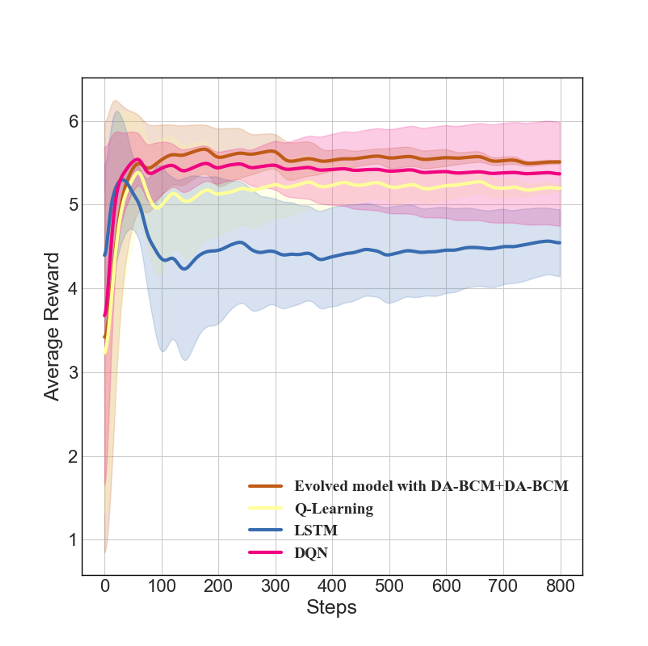}
\caption{Comparative Experimental results on Two Tasks. \textbf{a:} comparison results of four models in T-maze. \textbf{b:} comparison results of four models in Flappy Bird.}
\label{fig:lstm}
\end{figure*}
\section*{Discussion}
Evolution has not only designed the brain's general connectivity patterns but has also optimized a multi-scale plasticity coordinated learning rule, endowing the brain with the ability to flexibly adapt to reversal learning and enabling efficient online learning. Inspired by this, this paper proposed a structurally and functionally optimized LSM model that incorporates adaptive structural evolution and biologically plausible DA-BCM learning rule. Experimental results demonstrated that the structural evolution of the liquid layer and the DA-BCM regulation of the liquid layer and the readout layer significantly improved multiple decision-making tasks. 

Most existing works~\cite{34, 31, 64, 67,35,3,25,68,71} used backpropagation-based methods (which are suitable for hierarchical networks) to optimize the readout layer without considering the optimization of the liquid layer, or only adopted unsupervised STDP to optimize the liquid layer. Our model proposed a DA-BCM learning rule for both the liquid layer and the readout layer, which shows more biologically plausible. In addition, unlike existing structural search methods that directly search for the highest-performing structure, we took inspiration from the evolutionary mechanism and optimized the structure of the LSM according to its internal properties.
Here, we would like to compare our approach with other reinforcement learning models, including the classical Q-learning~\cite{watkins1992q} ,DQN\cite{mnih2013playing}, and LSTM~\cite{hochreiter1997long} (learning via policy gradient algorithm) with recurrent structure.

In LSTM configuration, the network consists of one layer of LSTM with $128$ hidden neurons and one fully connected layer. The Bellman equation Q-learning uses as Eq.\ref{ql}, where $\gamma=0.9$, $\alpha=0.1$. Agent's action is selected according to the $\epsilon$-greedy algorithm ($\epsilon=0.8$), which means that there is a probability of $0.2$ for each selection to explore the action space randomly. The reward discount value $\gamma$ and learning rate $\alpha$ are set to $0.99$ and $0.1$, respectively, in DQN. The loss function of the Q network is constructed in the form of mean square error, as shown in Eq.\ref{dqn}. The DQN network, which is fully connected, consists of three layers, the input layer, the hidden layer (with a size of 50), and the output layer. In Eq.\ref{dqn} $\gamma$ is set to 0.86. Learning rate in the used adam optimizer is set to 0.1.

For fairness, multiple experiments are performed for each comparison algorithm, and the performance is averaged. The results for LSTM, Q-learning, and DQN are averaged over multiple runs ($n=20$), where LSTM and DQN run 1000 episodes each.

\begin{align}\label{ql}
 &{}&Q(s, a)=Q(s, a)+\alpha\left[R(s, a)+\gamma \max _{a^{\prime}} Q^{\prime}\left(s^{\prime}, a^{\prime}\right)-Q(s, a)\right]
\end{align}
\begin{align}\label{dqn}
&{}&\omega^*=\arg \min _\omega \frac{1}{2 N} \sum_{i=1}^N\left[Q_\omega\left(s_i, a_i\right)-\left(r_i+\gamma \max _{a^{\prime}} Q_\omega\left(s_i^{\prime}, a^{\prime}\right)\right)\right]^2
\end{align}

Table \ref{tab:ctmaze} and Fig \ref{fig:lstm} compare the average reward of the evolved $N_{opt}$ individuals under different learning rule applications in detail. From the results, it can be seen that the efficiency of our proposed model is better than the comparison algorithms in terms of both mean and stability (variance). In fact, by combining \ref{tab:bird} and \ref{tab:bird}, it can be found that three evolutionary LSMs (DA-BCM+DA-BCM, STDP+DA-BCM, none+ DA-BCM) outperform LSTM and Q-learning in two tasks. We can also see that on the T-maze task, the performance of LSTM and DQN are significantly weaker than other models, and the variance of LSTM is very large, which may be caused by too many parameters that bring overfitting in a small sample learning task. In Flappy Bird, although DQN performance is better than LSTM and Q-Learning, the variance is very large. The overall efficiency is not as good as our model.

\begin{table}[h]
  \centering
  \caption{For different methods, the final performances (average reward R and its variance) on T-maze and Flappy Bird tasks.}
    \begin{tabular}{cccccc}
    \toprule
    \multicolumn{2}{c}{Learning Methods} & \multicolumn{2}{c}{T-maze} & \multicolumn{2}{c}{Flappy Bird} \\
    \midrule
    \multicolumn{2}{c}{Q-learning\cite{watkins1992q}} & \multicolumn{2}{c}{348.4$\pm$12.81} &   \multicolumn{2}{c}{5.19$\pm$0.04} \\
    \multicolumn{2}{c}{LSTM\cite{hochreiter1997long}} & \multicolumn{2}{c}{169.95$\pm$72.61} & \multicolumn{2}{c}{4.54$\pm$0.40} \\
              \multicolumn{2}{c}{DQN\cite{mnih2013playing}} & \multicolumn{2}{c}{80.1$\pm$11.68} & \multicolumn{2}{c}{5.36$\pm$0.62} \\
      \multicolumn{2}{c}{\textbf{Ours}} & \multicolumn{2}{c}{\textbf{464.4$\pm$5.15}} & \multicolumn{2}{c}{\textbf{5.43$\pm$0.03}} \\

    \bottomrule
    \end{tabular}%
  \label{tab:ctmaze}%
\end{table}%

\textbf{Computational Cost Analysis.} We also consider the impact of the computational cost of the model on fairness. Take T-maze for example. In our experiments, LSTM input size is 4, the hidden layer size is 128, and the total parameters are computed as Eq.\ref{eq34}:

\vspace{-0.5cm}
\begin{align}\label{eq34}
&{}&4*(4*128+128^2+128)=68096
\end{align}
 For Q-learning, only a state-action table of size 28*3=84 needs to be stored. For DQN, regardless of the Q table, the parameters of the three fully connected layers are $4*50*3=6000$. As for the LSM model we proposed, considering that the connection density of the evolved model is less than 2\%, the number of connections of the liquid layer is up to about $64*64*0.02=81.92$. Including the number of connections between the liquid layer and the input and output, the total number of parameters is about 109.92 on average, of the same magnitude as Q-learning. Therefore, the computational cost of our proposed model belongs to a low level compared to DQN and LSTM.

To sum up, this work breaks through the fixed deep hierarchical network structure that relies on BP optimization used in AI, and develops a multi-scale biological plasticity coordinated learning rule (instead of BP) and an efficient structure evolution for LSM. Because the proposed model borrows the information processing mechanism of the brain from the structure and function, it is more biologically plausible, more flexible and efficient, and naturally more suitable for developing human-like cognitive intelligence. Although this paper demonstrates the superiority of our proposed evolutionary LSM model in terms of model efficiency and computational cost, there are still some limitations. For example, in learning algorithms, there is still more room for exploration for developing brain-inspired models, and many neural mechanisms are waiting for us to investigate and fully apply in the AI field. This paper focuses on the small sample learning environment. Other application scenarios can also be used to further explore more energy-efficient self-organizing brain-inspired evolutionary spiking neural networks.

\section*{Methods}

\subsection*{LSM Architecture}

\subsubsection*{Leaky Integrate-and-Fire (LIF) Neuron Model}
Neuron model used in the proposed LSM is implemented with LIF model \cite{Lapicque1999L}, which can be simulated by Eq.\ref{eq3}:

\vspace{-0.5cm}
\begin{align}\label{eq3}
&{}&\tau_{\mathrm{m}} \frac{d V_{\mathrm{m}}(t)}{d t}=I(t)-V_{\mathrm{m}}(t)
\end{align}
$V_{m}$ is the voltage across the cell membrane, $I$ is the external input, and $\tau_{m}=2.0$ is the membrane potential time constant. The post-synaptic neuron voltage accumulates from $V_{reset}=0$ exponentially when a current spike is received from pre-synaptic neurons, producing an output spike as soon as the threshold $V_{th}=1.0$ is reached.

\subsubsection*{Liquid Layer Initialization}
In the experiment, the number of neurons in the liquid layer was set to 10*10, totaling 100. Inspired by neuron connections in the mammalian visual cortex, we set the connection probability $p$ between neurons $i$ and $j$ to exponentially inversely proportional to the Euclidean distance $d(i,j)$~\cite{53}. The closer the distance, the higher the connection probability, which is defined as the following:

\vspace{-0.5cm}
\begin{align}\label{eq7}
&{}&p=\left(e^{-\frac{1}{\lambda^{2}}}\right)^{d^{2}}
\end{align}

$\lambda$ is the parameter that controls the average number of synaptic connections and the average distance between neurons. To prevent neurons that are too far apart from connecting, a mask matrix $M_{dis}$ is added, combined with $p$ to form the weight matrix $W_l$ of the liquid layer as Eq.\ref{eq13}:

\vspace{-0.5cm}
\begin{align}\label{eq13}
&{}&W_{l}&=\alpha *M_{dis}*M_{sparse}*p
\end{align}

\begin{align}\label{eq13}
&{}&M^{i,j}_{dis}&=\begin{cases}
 1, d(i,j)<D_{th}
\\0, d(i,j)>D_{th} \quad or\quad i=j
\end{cases}
\end{align}

Both $M_{dis}$ and $M_{sparse}$ are binary matrices. $M_{dis}$ helps to form locally interconnected liquid layer structures (self-connection is not allowed). $M_{sparse}$ describes a sparse binary matrix in which only a randomly selected 1\% of the connections have the value $M_{sparse}$ equal to 1 (sparser liquid density is to facilitate subsequent evolution operations). $\alpha=4$ is a constant.

\subsubsection*{Readout Layer Initialization}
To construct an effective readout layer structure, preventing many inactive neurons from being read out, we formulate the  connection weight $W_{r}$ between the liquid layer and the readout layer according to the state matrix $S$ as follows:

\vspace{-0.5cm}
\begin{align}\label{eqreadout}
&{}&W_{r}&=\beta *M_{r}*S*w_{rand}
\end{align}

$w_{rand}$ indicates a rand weight matrix. Both $M_{r}$ and $S$ are binary matrices. When the readout layer has $N_r$ neurons, all the liquid layer neurons are randomly divided into $N_r$ classes, making each liquid layer neuron connected to only one readout layer neuron. The resulting mask matrix $M_{r}$ specifies which liquid layer neurons are connected to which readout layer neurons. 0 in the state matrix $S$ represents that the neuron did not fire, and a 1 represents it fired. Therefore, there is no connection between the non-firing liquid neurons and all the readout neurons. 
$\beta=4$ is a constant. The number of liquid neurons connected to each output or input neuron is set to 4.

\begin{algorithm}[ht]
    \caption {The evolution and learning process of LSM.}
    \label{alg1}
    \begin{algorithmic}
        \begin{footnotesize} 
        \renewcommand{\algorithmicrequire}{\textbf{Output:}}
    \renewcommand{\algorithmicensure}{\textbf{Initialize:}}
        \ENSURE{Population of LSM structures $pop$;}
        \REQUIRE{Evolved LSM structures;}\\
        \FOR {$generations$}
            \FOR {$individuals$ in $pop$}
                \STATE {$offsprings=mutate(individuals)$;}
                \STATE {Calculate the $F_{offs}$ of offsprings based on Eq.\ref{eq1};}
                \STATE {$newIndividual=select(offsprings,F_{offs})$;}
                \STATE {$pop[individuals]=newIndividual$;}
        \ENDFOR
        \STATE {Calculate the $F_{pop}$ of population based on Eq.\ref{eq1};}
                \STATE {$goodInd=select(pop,F_{pop},rate)$;}
                \IF {Current generation $g<G_{th}$} 
                    \STATE {Random generate individuals $newInd$;}
                    \STATE {$pop=goodInd+newInd$;}
                    \ELSE
                    \STATE{$pop=choose(pop,N_{opt})$}
                \ENDIF
    
        \ENDFOR
            
        \end{footnotesize}
    \end{algorithmic}
\end{algorithm}

\subsection*{Evolutionary LSM Structure}

\begin{figure*}[ht]
    \centering
    \includegraphics[width=14cm]{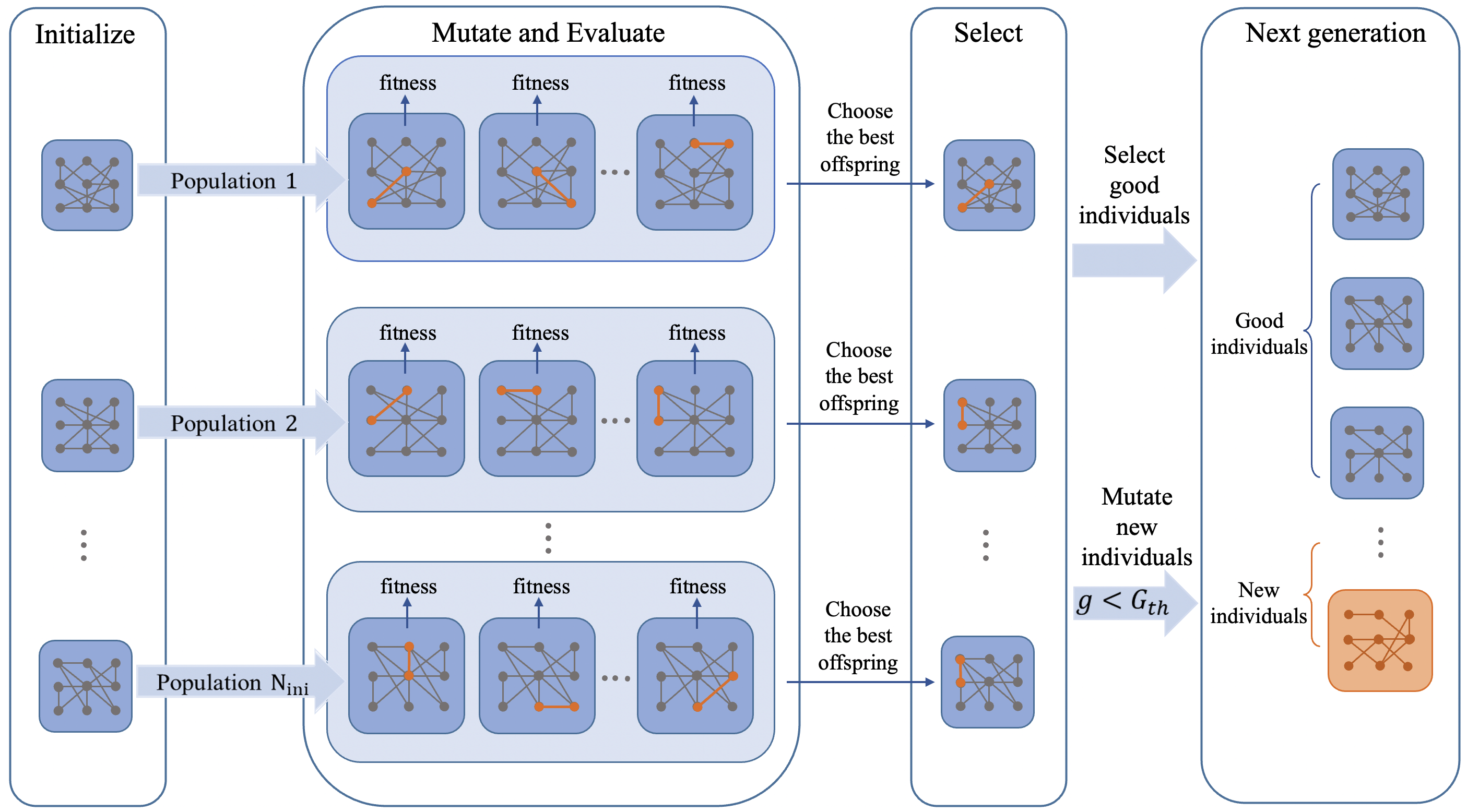}
    \caption{The procedure of evolutionary LSM structure.}
    \label{fig:EA}
\end{figure*}

First, we randomly initialized $N_{ini}$ LSMs to form a population $pop$ with their respective liquid layer connectivity matrices as chromosomes. For each chromosome, a population of offsprings is generated, and each offspring is mutated. According to the calculated fitness function, the optimal offspring corresponding to each chromosome is selected as a new chromosome. Meanwhile, to introduce innovation, the next generation consists of all optimal offsprings and partially regenerated random individuals. Evolution continues until the $N_{opt}$ individuals with the highest fitness in $G_{th}$ generation are selected as the output of the evolutionary algorithm. Fig.\ref{fig:EA} illustrates the detailed evolutionary process.

\subsubsection*{Initialization} 
We initialize $N_{ini}$ LSM individuals, each of which is an LSM structure. A chromosome is defined as a matrix representing an individual's liquid layer connectivity patterns:

\vspace{-0.5cm}
\begin{align}\label{eq1222}
&{}&Chrom^{i}&=\begin{cases}
 1, W_{l}^{i}>0
\\0, W_{l}^{i}=0,\quad 0\leq i \leq N_{ini}
\end{cases}
\end{align}
$i$ is the number of the individual, and $W_l$ is the liquid layer connection weight of the $i$th individual defined in Eq.\ref{eq13}.

\subsubsection*{Mutation} 

Each chromosome generates multiple offsprings (collectively a chromosome population) and mutates them: randomly select an inactive neuron (firing no spikes) among all liquid neurons and connect it with a surrounding active neuron (firing at least one spike).

\subsubsection*{Evaluation} 
The Separation Property (SP) was proposed by \cite{26} together with the concept of LSM as a measure of performance, which calculates the separation between the internal system state trajectories produced by two different input streams. There are many methods in current research to measure the SP of LSM, here we refer to \cite{28} to design an SP function to measure the quality of the liquid layer. We first calculate a state matrix $S$ (1 for fired, otherwise 0) of the liquid layer based on input and then compute the SP according to the following formula:

\vspace{-0.5cm}
\begin{align}\label{eq1}
    &{}&SP&=rank(S)
\end{align}

$rank(S)$ means the rank of matrix $S$. The larger the value, the stronger the separation property of LSM. After mutation, we calculate the separation property of offsprings obtained by mutation referring to Eq.\ref{eq1} as the fitness function.

\subsubsection*{Selection} 
Based on the fitness of all offsprings $F_{offs}$, select the one with the largest fitness in the chromosome population to replace the individual. In the first $G_{th}$ generation, the next generation consists of individuals with high fitness and new individuals explored randomly as a proportion of $rate$ of the entire population. After $G_{th}$ times of evolution, the new generation uses the experience of multiple iterative optimizations to select $N_{opt}$ individual with the highest fitness as the evolution output.

The algorithm process of evolving the LSM architecture is Algorithm.\ref{alg1}.

\subsection*{DA-BCM for Training Evolved LSM}

\begin{figure*}[ht]
    \centering
    \includegraphics[width=16cm]{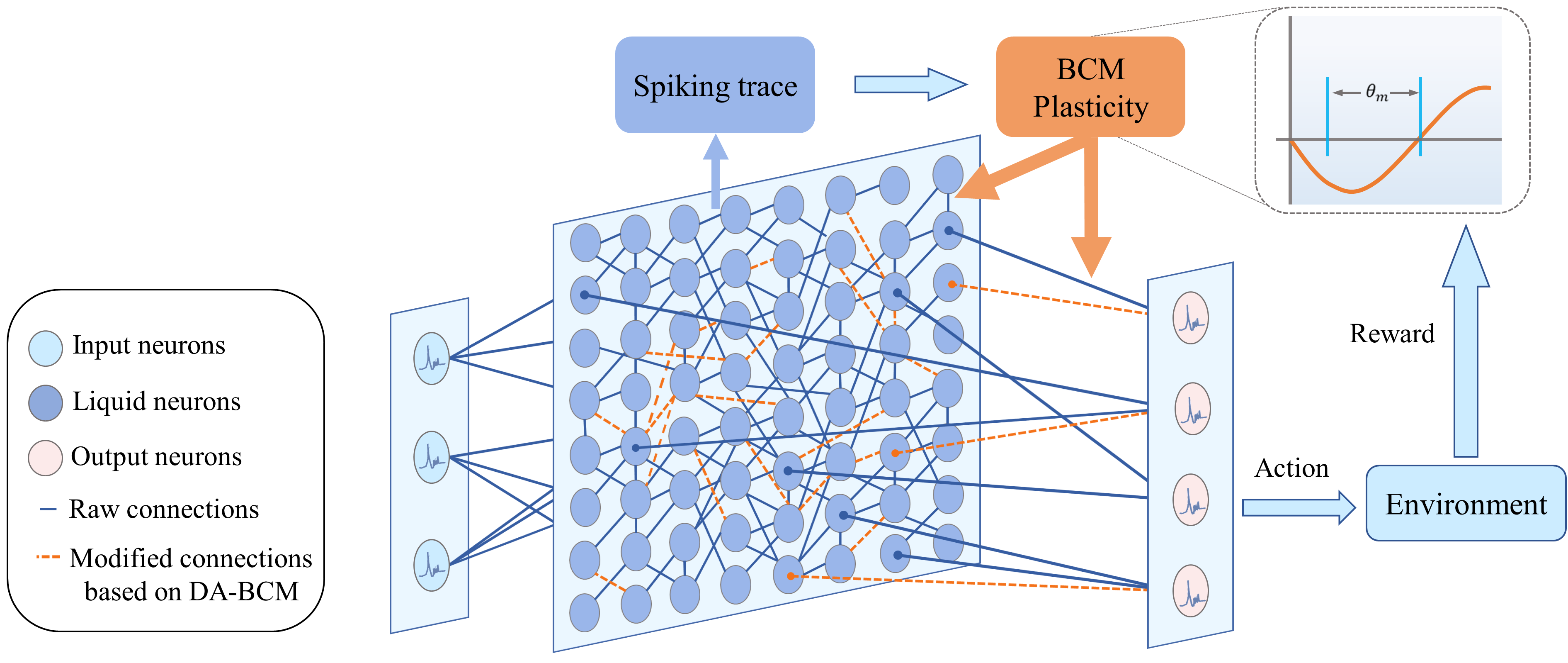}
    \caption{DA-BCM optimizes the synaptic weights of the evolved LSM.}
    \label{fig:da-bcm}
\end{figure*}

After evolving LSM, we incorporated multi-scale biological-inspired learning rules such as local synaptic plasticity and global dopamine regulation for optimizing the synaptic strength. As shown in Fig.\ref{fig:da-bcm}, the learning process updates the connection weights within the liquid layer and between the readout layers according to local BCM plasticity and global dopamine regulation.

\subsubsection*{Local BCM Synaptic Plasticity}
Since gradient-based learning rules are inconsistent with biological reality, we employed a more biologically plausible mechanism of synaptic plasticity: BCM rules \cite{10}, combined with dopamine global regulation to simulate the effects of reward and historical memory on behavior, encoding readout neurons target spatio-temporal dynamics. BCM was first used to explain how cortical neurons simultaneously undergo LTP or LTD depending on the different regulatory stimulation protocols applied to pre-synaptic neurons \cite{11}. According to BCM, the activity of the postsynaptic neuron strengthens the connection, and the activity experience of the postsynaptic neuron determines the dynamic correction of the threshold. The synaptic strength update rule for the activity of pre- and post-synaptic neurons is as follows:

\vspace{-0.5cm}
\begin{align}\label{eq8}
&{}&\frac{d m(t)}{d t}=\phi(e_{post}^{t}) e_{pre}^{t}-\epsilon m(t)
\end{align}

$m$ is the weight between pre- and post-synaptic neurons. $\epsilon$ is a coefficient that decays uniformly over time. $\phi$ is the BCM modification function that adjusts according to the neural spiking trace of the postsynaptic neuron, incorporating a sliding activity-dependent modification threshold $\theta_{m}$ to allow bidirectional synaptic modification. $e_{pre}^{t}$ is the spiking trace of the presynaptic neuron at time $t$ and $e_{post}^{t}$ is the spiking trace of the postsynaptic neuron at time $t$, which are calculated as:

\vspace{-0.5cm}
\begin{align}\label{eq100}
&{}&e_{pre}^{t}=\tau_{bcm} e_{pre}^{t-1}+o_{pre}^{t}
\end{align}

\begin{align}\label{eq100}
&{}&e_{post}^{t}=\tau_{bcm} e_{post}^{t-1}+o_{post}^{t}
\end{align}
Where $o_{pre}^{t}$ and $o_{post}^{t}$ denote the spikes of pre- and post-synaptic neurons, respectively. $\tau_{bcm}$ is the time decay constant. $\phi$ is defined as:

\begin{align}\label{eq101}
&{}&\phi(e)=e (e-\theta_{m})
\end{align}

The sliding threshold $\theta_{m}$ is dynamically updated according to the average value of the trace $e$ over a period of time.

\subsubsection*{Global Dopamine Regulation}

\cite{19} proposed the "reward prediction error hypothesis" that dopamine neurons encode reward and punishment signals during interacting with the environment. Related studies have introduced the learning rules of reward regulation into deep spiking neural networks\cite{mozafari2019bio} and  multi-brain regions coordinated SNNs\cite{zhao2020neural,zhao2018brain}. Further, 
reward-modulated STDP\cite{Eugene2007,Hongjian2021,Nicolas2016Neuromodulated} integrates dopamine regulation and STDP could solve the problem of credit assignment in order to obtain more reward.

Here, inspired by the neural mechanisms of dopamine regulation, we propose a DA-BCM learning rule that integrates long-term dopamine regulation and local BCM synaptic plasticity. When receiving an external reward signal, dopamine forms a global long-term regulation, combining with BCM plasticity to adaptively adjust synaptic strengthen for the liquid layer and readout layer. The DA-BCM learning rule is as follows:

\vspace{-0.5cm}
\begin{align}\label{eq12}
&{}&\frac{d m(t)}{d t}=DA*(\phi(e_{post}^{t}) e_{pre}^{t}-\epsilon m(t))
\end{align}
Here, $DA$ stands for dopamine signal.

\bibliography{sample}

\section*{Acknowledgements}
This work is supported by the National Key Research and Development Program (Grant No. 2020AAA0107800), the Strategic Priority Research Program of the Chinese Academy of Sciences (Grant No. XDB32070100), the National Natural Science Foundation of China (Grant No. 62106261). 

\section*{Code availability}

All original code has been deposited at \href{https://github.com/Agnes233/evolve_LSM}{https://github.com/Agnes233/evolveLSM} and is publicly available as of the date of publication
\section*{Author contributions statement}

W.P.,F.Z. and Y.Z. designed the study. W.P.,F.Z. and B.H.performed the experiments and the analyses. W.P.,F.Z. and Y.Z. wrote the paper.
\section*{Competing interests}
The authors declare no competing interests.

\end{document}